\documentclass[conference]{IEEEtran}
\pdfoutput=1
\IEEEoverridecommandlockouts
\usepackage{amsmath,amssymb,amsfonts}
\usepackage{algorithmic}

\makeatletter
\let\MYcaption\@makecaption
\makeatother
\usepackage[font=footnotesize]{subcaption}
\makeatletter
\let\@makecaption\MYcaption
\makeatother

\usepackage{graphicx}
\usepackage{textcomp}
\usepackage{xcolor}
\usepackage[style=ieee,
            doi=false,
            url=false,
            mincitenames=1,
            maxcitenames=1,
            natbib=true,
            minbibnames=6,
            maxbibnames=6,
            backend=biber]{biblatex}
\AtBeginBibliography{\small}
\usepackage{longtable}
\usepackage{acro}

\DeclareAcronym{2D}{
    short = 2D,
    long = two-dimensional,
}
\DeclareAcronym{3D}{
    short = 3D,
    long = three-dimensional,
}
\DeclareAcronym{AD}{
    short = AD,
    long = automated driving,
}
\DeclareAcronym{ADAM}{
    short = ADAM,
    long = Adaptive Moment Estimation,
}
\DeclareAcronym{AI}{
    short = AI,
    long = artifical intelligence,
}
\DeclareAcronym{ANN}{
    short = ANN,
    long = artificial neural network,
}
\DeclareAcronym{AOE}{
    short = AOE,
    long = average orientation error,
}
\DeclareAcronym{AORE}{
    short = AORE,
    long = average odometry rotation error,
}
\DeclareAcronym{AOTE}{
    short = AOTE,
    long = average odometry translation error,
}
\DeclareAcronym{AP}{
    short = AP,
    long = average precision,
}
\DeclareAcronym{AR}{
    short = AR,
    long = average recall,
}
\DeclareAcronym{ARE}{
    short = ARE,
    long = average rotation error,
}
\DeclareAcronym{ASE}{
    short = ASE,
    long = average scale error,
}
\DeclareAcronym{ATE}{
    short = ATE,
    long = average translation error,
}
\DeclareAcronym{AVE}{
    short = AVE,
    long = average velocity error,
}
\DeclareAcronym{BBA}{
    short = BBA,
    long = basic belief assignment,
    short-plural-form = BBAs,
    long-plural-form = basic belief assignments
}
\DeclareAcronym{BCE}{
    short = BCE,
    long = binary cross entropy,
}
\DeclareAcronym{BiFPN}{
    short = BiFPN,
    long = Bidirectional Feature Pyramid Network,
}
\DeclareAcronym{BN}{
    short = BN,
    long = batch normalization,
}
\DeclareAcronym{CCL}{
    short = CCL,
    long = connected-components labeling,
}
\DeclareAcronym{CE}{
    short = CE,
    long = cross entropy,
}
\DeclareAcronym{CNN}{
    short = CNN,
    long = convolutional neural network,
}
\DeclareAcronym{DAG}{
    short = DAG,
    long = directed acyclic graph,
}
\DeclareAcronym{DBSCAN}{
    short = DBSCAN,
    long = density-based spatial clustering of applications with noise,
}
\DeclareAcronym{DEVN}{
    short = DEVN,
    long = directed evidential network with conditional belief functions,
}
\DeclareAcronym{DL}{
    short = DL,
    long = deep learning,
}
\DeclareAcronym{eIoU}{
    short = eIoU,
    long = evidential intersection over union,
}
\DeclareAcronym{ENC}{
    short = ENC,
    long = evidential network with conditional belief functions,
}
\DeclareAcronym{FLOP}{
    short = FLOP,
    long = floating-point operation,
}
\DeclareAcronym{FOD}{
    short = FoD,
    long = frame of discernment,
    long-plural-form = frames of discernment
}
\DeclareAcronym{FOV}{
    short = FOV,
    long = field of view,
}
\DeclareAcronym{FPN}{
    short = FPN,
    long = Feature Pyramid Network,
}
\DeclareAcronym{FPS}{
    short = FPS,
    long = frames per seconds,
}
\DeclareAcronym{GNC}{
    short = GNC,
    long = Graduated Non-Convexity,
}
\DeclareAcronym{GMC}{
    short = GMC,
    long = Geman McClure,
}
\DeclareAcronym{GPS}{
    short = GPS,
    long = Global Positioning System,
}
\DeclareAcronym{GPU}{
    short = GPU,
    long = Graphics Processing Unit,
}
\DeclareAcronym{IMU}{
    short = IMU,
    long = Inertial Measurement Unit,
}
\DeclareAcronym{IoU}{
    short = IoU,
    long = intersection over union,
}
\DeclareAcronym{KLD}{
    short = KLD,
    long = Kullback-Leiber divergence,
}
\DeclareAcronym{lidar}{
    short = LiDAR,
    long = light detection and ranging,
}
\DeclareAcronym{LLS}{
    short = LLS,
    long = linear Least-Squares,
}
\DeclareAcronym{LS}{
    short = LS,
    long = Least-Squares,
}
\DeclareAcronym{mAOE}{
    short = mAOE,
    long = mean average orientation error,
}
\DeclareAcronym{mAORE}{
    short = mAORE,
    long = mean average odometry rotation error,
}
\DeclareAcronym{mAOTE}{
    short = mAOTE,
    long = mean average odometry translation error,
}
\DeclareAcronym{mAP}{
    short = mAP,
    long = mean average precision,
}
\DeclareAcronym{mARE}{
    short = mARE,
    long = mean average rotation error,
}
\DeclareAcronym{mASE}{
    short = mASE,
    long = mean average scale error,
}
\DeclareAcronym{mATE}{
    short = mATE,
    long = mean average translation error,
}
\DeclareAcronym{mAVE}{
    short = mAVE,
    long = mean average velocity error,
}
\DeclareAcronym{MIB}{
    short = MIB,
    long = multi-instance Bernoulli,
}
\DeclareAcronym{ML}{
    short = ML,
    long = machine learning,
}
\DeclareAcronym{NMS}{
    short = NMS,
    long = non-maximum suppression,
}
\DeclareAcronym{OLS}{
    short = OLS,
    long = ordinary Least-Squares,
}
\DeclareAcronym{PCR}{
    short = PCR,
    long = partial conflict redistribution
}
\DeclareAcronym{PDF}{
    short = PDF,
    long = probability density function,
}
\DeclareAcronym{PHD}{
    short = PHD,
    long = probability hypothesis density,
}
\DeclareAcronym{PHD/MIB}{
    short = PHD/MIB,
    long = probability hypothesis density / multi-instance Bernoulli,
}
\DeclareAcronym{radar}{
    short = RaDAR,
    long = radio detection and ranging,
}
\DeclareAcronym{RFS}{
    short = RFS,
    long = random finite set,
}
\DeclareAcronym{RE}{
    short = RE,
    long = rotational error,
}
\DeclareAcronym{ReLU}{
    short = ReLU,
    long = rectified linear unit,
}
\DeclareAcronym{RPN}{
    short = RPN,
    long = region proposal network,
}
\DeclareAcronym{SGD}{
    short = SGD,
    long = stochastic gradient descent,
}
\DeclareAcronym{sonar}{
    short = Sonar,
    long = Sound Navigation and Ranging,
}
\DeclareAcronym{TE}{
    short = TE,
    long = translational error,
}
\DeclareAcronym{TLS}{
    short = TLS,
    long = truncated Least-Squares,
}
\DeclareAcronym{UBS}{
    short = UBS,
    long = uniform B-spline,
}
\DeclareAcronym{ER}{
    short = ER,
    long = evidential reasoning,
}
\DeclareAcronym{WLS}{
    short = WLS,
    long = weighted Least-Squares,
}

\usepackage{glossaries}

\newglossaryentry{sensorgm}
{
  name=sensor grid map,
  description={sensor grid map}
}
\newglossaryentry{fusedgm}
{
  name=fused grid map,
  description={fused grid map}
}
\newglossaryentry{cam}
{
  name=Camera,
  description={Camera }
}
\newglossaryentry{kitti360}
{
  name=Kitti-360,
  description={Kitti-360 }
}
\newglossaryentry{monocam}
{
  name=monocular camera,
  description={monocular camera }
}
\newglossaryentry{stereocam}
{
  name=stereo camera,
  description={stereo camera }
}
\newglossaryentry{occdyn}
{
  name=occupancy dynamics,
  description={occupancy dynamics}
}
\newglossaryentry{occsem}
{
  name=occupancy semantics,
  description={occupancy semantics}
}

\usepackage[pdfstartview=XYZ,
bookmarks=true,
colorlinks=true,
linkcolor=black,
urlcolor=black,
citecolor=black,
bookmarks=true,
linktocpage=true, % makes the page number as hyperlink in table of content
hyperindex=true
]{hyperref}
\usepackage[capitalize]{cleveref}
\usepackage{siunitx}

\usepackage{amsfonts, amsmath, amssymb, amsthm}
\usepackage{mathdots}
\usepackage{mathtools}

\usepackage{bm}

\theoremstyle{definition}
\newcommand{\func}[1]{\mathrm{#1}}

\DeclarePairedDelimiterX{\inp}[2]{\langle}{\rangle}{#1, #2}

\newcommand{\denge}[1]{\mathrm{e}_\mathrm{D}(#1)}
\newcommand{\dengns}[1]{\mathrm{ns}_\mathrm{D}(#1)}
\newcommand{\dengd}[1]{\mathrm{d}_\mathrm{D}(#1)}

\newcommand{\real}{\mathbb{R}}

\newcommand{\bba}[1]{\func{m}(#1)}

\newcommand{\powerset}[1]{\mathcal{P}(#1)}

\newcommand{\frameo}{\Omega_o}

\newcommand{\cl}{\mathcal{S}}

\newcommand{\rect}{\mathcal{R}}

\newcommand{\grid}{\mathcal{G}}

\newcommand{\cell}{C}

\newcommand{\gt}{\mathrm{ref}}

\newcommand{\eiou}{\mathrm{eIoU}}

\newcommand{\etp}{\mathrm{eTP}}
\newcommand{\efp}{\mathrm{eFP}}
\newcommand{\efn}{\mathrm{eFN}}
\newcommand{\fbba}{\mathrm{m}}

\newcommand{\gm}{\mathrm{g}}

\newcommand{\pr}[1]{\mathrm{Pr}(#1)}

\newcommand{\defeq}{\vcentcolon=}

\makeatletter
\newcommand{\ocap}{\mathbin{\mathpalette\make@circled\smallcap}}
\newcommand{\make@circled}[2]{%
  \ooalign{$\m@th#1\smallbigcirc{#1}$\cr\hidewidth$\m@th#1#2$\hidewidth\cr}%
}
\newcommand{\smallbigcirc}[1]{%
  \vcenter{\hbox{\scalebox{0.77778}{$\m@th#1\bigcirc$}}}%
}
\newcommand{\smallcap}{%
  \vcenter{\hbox{\scalebox{0.6}{$\cap$}}}%
}
\makeatother

\newcommand{\lidarkitti}{Velodyne HDL-64E}
\newcommand{\kittigeiger}{Kitti-360}

\usepackage{tikz}
\usepackage{ifthen}

\usepackage[inkscapepath=out/svg-inkscape/, notransparent]{svg}
\svgpath{figures/}

\graphicspath{{figures/},{tikz/}}

\usepackage{pgfplots}
\usepackage{stackengine}
\usepackage{tikzexternal}
\usepgfplotslibrary{
    colorbrewer,
    groupplots,
}
\usetikzlibrary{patterns}

\pgfplotsset{
    bar cycle list/.style={
            cycle list name=Set1,
            every axis plot/.append style={fill, thin},
        },
    compat=1.15,
    colormap/Spectral,
    colormap={reverse Spectral}{
            indices of colormap={
                    \pgfplotscolormaplastindexof{Spectral},...,0 of Spectral}
        },
    cycle list/Set1,
    cycle list name=Set1,
    enlarge x limits=0.05,
    enlarge y limits=0.05,
    every axis/.append style={semithick},
    every axis plot/.append style={thick},
    grid style={
            opacity=0.5,
        },
    height=0.25\linewidth,
    scale only axis=true,
    scaled x ticks=false,
    scaled y ticks=false,
    title style={
            anchor=north east,
            at={(0.99, 0.95)},
            fill=white,
            fill opacity=0.5,
            text opacity=1,
        },
    width=0.9\linewidth,
}

\newrobustcmd*{\tribetter}{\tikz{\filldraw[draw=black!30!green,fill=black!30!green] (0,0) -- (1mm,0) -- (0.5mm,1mm);}}
\newrobustcmd*{\triworse}{\tikz{\filldraw[draw=black!30!red,fill=black!30!red] (0,0) -- (1mm,0) -- (0.5mm,1mm);}}

\pgfkeys{/entities/.cd,
  file/.initial=,
  angle/.initial=0,
  width/.initial=,
  plot_ego/.initial=1,
  plot_hsv/.initial=1,
  roi_x_min/.initial=-50,
  roi_y_min/.initial=-50,
  roi_x_max/.initial=50,
  roi_y_max/.initial=50,
  hsv_xmin/.initial=40,
  hsv_xmax/.initial=46,
  hsv_ymin/.initial=-46,
  hsv_ymax/.initial=-40,
  }
\def\set@keys#1{%%
  \pgfkeys{/entities/.cd,#1}}
\def\get#1{%%
  \pgfkeysvalueof{/entities/#1}}
\newcommand{\quadgrid}[1]{
\begin{tikzpicture}
    \set@keys{#1}%%
    \begin{axis}[
            axis on top,
            major grid style={very thin},
            enlargelimits=false,
            grid=major,
            height=\get{width}\linewidth,
            width=\get{width}\linewidth,
            ticklabel style = {font=\tiny}
        ]
        \addplot[plot graphics/node/.append style={yscale=-1,anchor=north west}] graphics[xmin=\get{roi_x_min},xmax=\get{roi_x_max},ymin=\get{roi_y_min},ymax=\get{roi_y_max}, includegraphics={trim=0 0  0 0,clip,keepaspectratio}] {\get{file}};
        \addplot[rotate around={\get{angle}:(0,0)}] graphics[xmin=-2,xmax=3.2,ymin=-1.3,ymax=1.3] {bertha/car};
    \end{axis}
\end{tikzpicture}
}

\newcommand{\quadgridhsv}[1]{
\begin{tikzpicture}
    \set@keys{#1}%%
    \begin{axis}[
            axis on top,
            major grid style={very thin},
            enlargelimits=false,
            grid=major,
            width=\get{width}\linewidth,
            height=(\get{roi_y_max}-\get{roi_y_min})/(\get{roi_x_max}-\get{roi_x_min})*\get{width}\linewidth,
            axis equal,
            ticklabel style = {font=\tiny}
        ]
        \addplot[plot graphics/node/.append style={yscale=-1,anchor=north west}] graphics[xmin=\get{roi_x_min},xmax=\get{roi_x_max},ymin=\get{roi_y_min},ymax=\get{roi_y_max}, includegraphics={trim=0 0  0 0,clip,keepaspectratio}] {\get{file}};
        \ifthenelse{\equal{\get{plot_ego}}{1}}
           {\addplot[rotate around={\get{angle}:(0,0)}] graphics[xmin=-2,xmax=3.2,ymin=-1.3,ymax=1.3] {bertha/car};}{}
        \ifthenelse{\equal{\get{plot_hsv}}{1}}
            {\addplot graphics[xmin=\get{hsv_xmin},xmax=\get{hsv_xmax},ymin=\get{hsv_ymin},ymax=\get{hsv_ymax}] {hsv_wheel};}{}
    \end{axis}
\end{tikzpicture}
}

\pgfkeys{/wg/entities/.cd,
  file/.initial=,
  width/.initial=,
  }
\def\wgset@keys#1{%%
  \pgfkeys{/wg/entities/.cd,#1}}
\def\wgget#1{%%
  \pgfkeysvalueof{/wg/entities/#1}}
\newcommand{\widegrid}[1]{
\begin{tikzpicture}
    \wgset@keys{#1}%%
    \begin{axis}[
            axis on top,
            major grid style={very thin},
            enlargelimits=false,
            grid=major,
            height=0.4*\wgget{width}\linewidth,
            width=\wgget{width}\linewidth,
            xtick={-40, -20, ..., 40.0001},
            ytick={-40, -20, ..., 40.0001},
            xticklabels={-40, -20, ..., 40},
            yticklabels={-40, -20, ..., 40},
            ticklabel style = {font=\tiny}
        ]
        \addplot[plot graphics/node/.append style={yscale=-1,anchor=north west}] graphics[xmin=-50,xmax=50,ymin=-20,ymax=20, includegraphics={trim=0 300  0 300,clip,keepaspectratio}] {\wgget{file}};
        \addplot graphics[xmin=-2,xmax=3.2,ymin=-1.3,ymax=1.3] {bertha/car};
    \end{axis}
\end{tikzpicture}
}

\newcommand{\widegridcam}[1]{
\begin{tikzpicture}
    \wgset@keys{#1}%%
    \begin{axis}[
            axis on top,
            major grid style={very thin},
            enlargelimits=false,
            grid=major,
            height=0.67*\wgget{width}\linewidth,
            width=\wgget{width}\linewidth,
            xtick={-10, 0, ..., 40.0001},
            ytick={-40, -20, ..., 40.0001},
            xticklabels={-10, 0, ..., 40},
            yticklabels={-40, -20, ..., 40},
            ticklabel style = {font=\tiny}
        ]
        \addplot[plot graphics/node/.append style={yscale=-1,anchor=north west}] graphics[xmin=-10,xmax=50,ymin=-20,ymax=20, includegraphics={trim=400 300 0 300,clip,keepaspectratio}] {\wgget{file}};
        \addplot graphics[xmin=-2,xmax=3.2,ymin=-1.3,ymax=1.3] {bertha/car};
    \end{axis}
\end{tikzpicture}
}

\newcommand{\widegridfront}[1]{
\begin{tikzpicture}
    \wgset@keys{#1}%%
    \begin{axis}[
            axis on top,
            major grid style={very thin},
            enlargelimits=false,
            grid=major,
            height=0.5*\wgget{width}\linewidth,
            width=\wgget{width}\linewidth,
            xtick={-40, -20, ..., 40.0001},
            xticklabels={-40, -20, ..., 40},
            ytick={-15, 0, 15.0001},
            yticklabels={-15, 0, 15},
            ticklabel style = {font=\tiny}
        ]
        \addplot[plot graphics/node/.append style={yscale=-1,anchor=north west}] graphics[xmin=-10,xmax=50,ymin=-15,ymax=15, includegraphics={trim=400 350  0 350,clip,keepaspectratio}] {\wgget{file}};
        \addplot graphics[xmin=-2,xmax=3.2,ymin=-1.3,ymax=1.3] {bertha/car};
    \end{axis}
\end{tikzpicture}
}

\newcommand{\widegridur}[1]{
\begin{tikzpicture}
    \wgset@keys{#1}%%
    \begin{axis}[
            axis on top,
            major grid style={very thin},
            enlargelimits=false,
            grid=major,
            height=2.5*\wgget{width}\linewidth,
            width=\wgget{width}\linewidth,
            xtick={-40, -20, ..., 40.0001},
            ytick={-40, -20, ..., 40.0001},
            xticklabels={-40, -20, ..., 40},
            yticklabels={-40, -20, ..., 40},
            ticklabel style = {font=\tiny}
        ]
        \addplot[plot graphics/node/.append style={yscale=-1,anchor=north west}]  graphics[xmin=-20,xmax=20,ymin=-50,ymax=50, includegraphics={trim=0 300  0 300,clip,keepaspectratio,angle=-90,origin=c}] {\wgget{file}};
        \addplot[rotate around={90:(0,0)}] graphics[xmin=-2,xmax=3.2,ymin=-1.3,ymax=1.3] {bertha/car};
    \end{axis}
\end{tikzpicture}
}

\newcommand{\widegridhsv}[1]{
\begin{tikzpicture}
    \wgset@keys{#1}%%
    \begin{axis}[
            axis on top,
            major grid style={very thin},
            enlargelimits=false,
            grid=major,
            height=0.4*\wgget{width}\linewidth,
            width=\wgget{width}\linewidth,
            xtick={-40, -20, ..., 40.0001},
            ytick={-40, -20, ..., 40.0001},
            xticklabels={-40, -20, ..., 40},
            yticklabels={-40, -20, ..., 40},
            ticklabel style = {font=\tiny}
        ]
        \addplot[plot graphics/node/.append style={yscale=-1,anchor=north west}] graphics[xmin=-50,xmax=50,ymin=-20,ymax=20, includegraphics={trim=0 300  0 300,clip,keepaspectratio}] {\wgget{file}};
        \addplot graphics[xmin=-2,xmax=3.2,ymin=-1.3,ymax=1.3] {bertha/car};
        \addplot graphics[xmin=42,xmax=48,ymin=-18,ymax=-12] {hsv_wheel};
    \end{axis}
\end{tikzpicture}
}

\newcommand{\widegridhsvur}[1]{
\begin{tikzpicture}
    \wgset@keys{#1}%%
    \begin{axis}[
            axis on top,
            major grid style={very thin},
            enlargelimits=false,
            grid=major,
            height=2.5*\wgget{width}\linewidth,
            width=\wgget{width}\linewidth,
            xtick={-40, -20, ..., 40.0001},
            ytick={-40, -20, ..., 40.0001},
            xticklabels={-40, -20, ..., 40},
            yticklabels={-40, -20, ..., 40},
            ticklabel style = {font=\tiny}
        ]
        \addplot[plot graphics/node/.append style={yscale=-1,anchor=north west}] graphics[xmin=-20,xmax=20,ymin=-50,ymax=50, includegraphics={trim=0 300  0 300,clip,keepaspectratio,angle=-90,origin=c}] {\wgget{file}};
        \addplot[rotate around={90:(0,0)}] graphics[xmin=-2,xmax=3.2,ymin=-1.3,ymax=1.3] {bertha/car};
        \addplot[rotate around={90:(15,-45)}] graphics[xmin=12,xmax=18,ymin=-48,ymax=-42] {hsv_wheel};
    \end{axis}
\end{tikzpicture}
}

\pgfkeys{/map/entities/.cd,
  min/.initial=0,
  max/.initial=1,
  width/.initial=1,
  }
\def\mapset@keys#1{%%
  \pgfkeys{/map/entities/.cd,#1}}
\def\mapget#1{%%
  \pgfkeysvalueof{/map/entities/#1}}
\newcommand{\colorbar}[1]{
\begin{tikzpicture}
    \fontsize{6pt}{9pt}\selectfont
    \mapset@keys{#1}%%
    \pgfplotscolorbardrawstandalone[ 
        colormap/jet,
        colorbar horizontal,
        point meta min=\mapget{min},
        point meta max=\mapget{max},
        colorbar style={
            width=\mapget{width}\linewidth,
            height=5pt,
            draw=white
        }
    ]
\end{tikzpicture}
}

\definecolor{colorstreet}{RGB}{128, 64, 128}
\definecolor{colorsidewalk}{RGB}{244, 35, 232}
\definecolor{colorterrain}{RGB}{152, 251, 152}
\definecolor{colorcar}{RGB}{0, 0, 142}
\definecolor{colorcyclist}{RGB}{119, 11, 32}
\definecolor{colorped}{RGB}{220, 20, 60}
\definecolor{colormov}{RGB}{0, 80, 100}
\definecolor{colornonov}{RGB}{102, 102, 156}
\definecolor{colorocc}{RGB}{0, 0, 0}
\definecolor{colorfree}{RGB}{240, 240, 240}

\newcommand{\colorbarev}{
    \centering
    \begin{tikzpicture}
        \fontsize{6pt}{9pt}\selectfont
        \coordinate (p);
        \foreach \name/\cl/\tcl in {
            Car/colorcar/white,
            Two-wheeler/colorcyclist/white,
            Pedestrian/colorped/white,
            Immobile/colornonov/white,
            Occupied/colorocc/white,
            Free/colorfree/black}
        {
          \node[rectangle, text=\tcl, fill=\cl, draw=white, minimum height=10pt, minimum width=20pt, anchor=west]
          (n) at (p) {\name};
          \coordinate (p) at ([xshift=-1pt]n.east);
        }
    \end{tikzpicture}
}

\newcommand{\colorbarevstack}{
    \fontsize{6pt}{9pt}\selectfont
    \coordinate (p) at (-2.95, -3.38);
    \node[rectangle, yshift=2.45cm, text=white, fill=colorcar, fill opacity=0.9, minimum height=10pt, minimum width=53pt, anchor=north]
    (n) at (p) {Car};
    \coordinate (p) at (n.south);
    \foreach \name/\cl/\tcl in {
        Two-wheeler/colorcyclist/white,
        Pedestrian/colorped/white,
        Immobile/colornonov/white,
        Occupied/colorocc/white,
        Free/colorfree/black}
    {
        \node[rectangle, text=\tcl, fill=\cl, fill opacity=0.9, minimum height=10pt, minimum width=53pt, anchor=north]
        (n) at (p) {\name};
        \coordinate (p) at (n.south);
    }
}

\definecolor{colorobjbl}{RGB}{0, 255, 0}
\definecolor{colorobjnbl}{RGB}{255, 0, 0}
\definecolor{colorgroundbl}{RGB}{0, 0, 255}
\definecolor{colorgroundnbl}{RGB}{0, 0, 0}

\definecolor{colorblack}{RGB}{0, 0, 0}
\definecolor{colorgray}{RGB}{150, 150, 150}
\definecolor{colorred}{RGB}{205, 34, 34}
\definecolor{colorgreen}{RGB}{63, 205, 34}
\definecolor{colorblue}{RGB}{34, 63, 205}

\begin{document}

\title{Sensor Data Fusion in Top-View Grid Maps using Evidential Reasoning with Advanced Conflict Resolution}

\author{Sven Richter\(^{1,\ast}\), Frank Bieder\(^{1,2}\), Sascha Wirges\(^{3}\) and
		Christoph Stiller\(^{1,2}\)%
\thanks{\(^1\) Authors are with the Institute of Measurement and Control Systems, Karlsruhe Institute of Technology (KIT), Karlsruhe, Germany.%
{\tt\small \{sven.richter, frank.bieder, stiller\}@kit.edu}}
\thanks{\(^2\) Authors are with the Mobile Perception Systems Group, FZI Research Center for Information Technology, Karlsruhe, Germany.}%
\thanks{\(^3\) Author is with the Bosch Center for Artifical Intelligence, Renningen, Germany.
{\tt\small sascha.wirges@de.bosch.com}}
\thanks{\(^\ast\) Corresponding author.}
}

\maketitle

% \IEEEpubid{\begin{minipage}{\textwidth}~\\[12pt] \centering%
% 	% 10.1000/xyz123~ % Insert your DOI after publication
% 	\copyright~2022 IEEE. Personal use of this material is permitted. Permission from IEEE must be obtained for all other uses, including reprinting/republishing this material for advertising or promotional purposes, collecting new collected works for resale or redistribution to servers or lists, or reuse of any copyrighted component of this work in other works.
%   \end{minipage}}
\IEEEpubid{\begin{minipage}{\textwidth}~\\[12pt] \centering%
	% 10.1000/xyz123~ % Insert your DOI after publication
	This work has been submitted to the IEEE for possible publication. Copyright may be transferred without notice, after which this version may no longer be accessible.
  \end{minipage}}
  \IEEEpubidadjcol

\begin{abstract}
    We present a new method to combine evidential top-view grid maps estimated based on heterogeneous sensor sources.
    Dempster's combination rule that is usually applied in this context provides undesired results with highly conflicting inputs.
    Therefore, we use more advanced \acl{ER} techniques and improve the conflict resolution by modeling the reliability of the evidence sources.
    We propose a data-driven reliability estimation to optimize the fusion quality using the \kittigeiger{} dataset.
    We apply the proposed method to the fusion of \acs{lidar} and \gls{stereocam} data and evaluate the results qualitatively and quantitatively.
    The results demonstrate that our proposed method robustly combines measurements from heterogeneous sensors and successfully resolves sensor conflicts.
\end{abstract}


\section{Introduction}

\begin{figure}[!t]
    % \centering
    \fontsize{8pt}{8pt}\selectfont
    \begin{tikzpicture}
        \node at (0,0) {\includesvg[width=\linewidth]{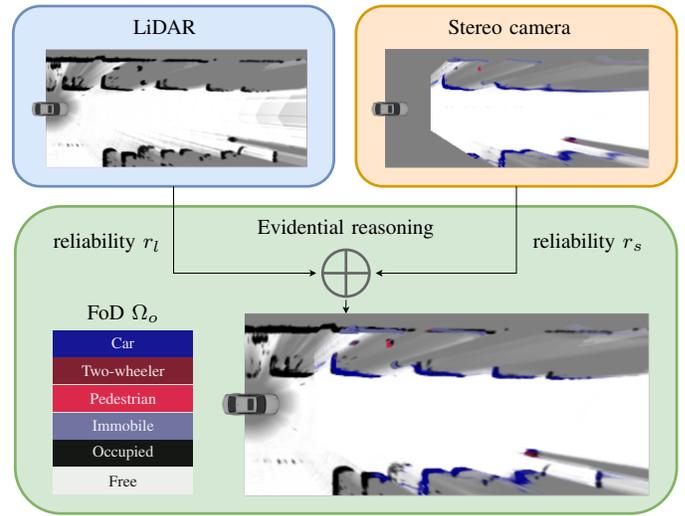}};
        \colorbarevstack
    \end{tikzpicture}
    \caption{The sensor data fusion framework presented in this work.
            Evidential grid maps modeling free space and semantic occupancy from \ac{lidar} and \gls{stereocam} are combined with advanced conflict handling using sensor reliability coefficients $r_l$ and $r_s$.}
    \label{fig:overview}
\end{figure}

For the navigation of automated vehicles, a detailed reconstruction of the environment is needed.
In order to benefit from the individual strengths of different sensing principles, heterogeneous sensor setups are used.
Top-view grid maps are especially suitable to model measurements from multimodal sensors and to combine them into one common representation.
They enable modeling a wide range of information such as free space, the presence of obstacles and their semantic state and simplify information fusion due to their structured data format.
Information fusion is particularly challenging, if contradicting information is combined.
The combination rules that are frequently used for combining evidence measures in top-view grid maps such as Dempster's or Yager's rule perform unintuitively in those cases.
We propose an adaptive method to handle this based on the \ac{ER} framework presented by \citet{Yang2013}.
In case the sources provide conjunctive estimates, it reduces to Dempster's rule.
If the sources provide conflicting estimates, the combination is adapted based on a reliability coefficient assigned to each source of evidence individually.
In this work, we combine evidential top-view grid maps estimated based on \ac{lidar} and \gls{stereocam} measurements.
The fusion framework is sketched in \Cref{fig:overview}.
The reliability coefficients $r_l$ for the \ac{lidar} and $r_s$ for the \gls{stereocam} are estimated to optimize the fusion performance with respect to the ground truth bounding primitives in the \kittigeiger{} dataset \cite{Liao2021ARXIV}.
\IEEEpubidadjcol
\section{Related Work}

When combining degrees of evidence from independent sensor sources in grid maps, Dempster's rule is usually applied \cite{Nuss2014,Tanzmeister2017}.
However, because of the above-mentioned shortcomings of this rule, other combination operators haven been explored as well:
\citet{Moras2015} applied the \ac{PCR}6 to the fusion of evidential occupancy grid maps.
They tested their method with simulated \ac{lidar} data that they fuse over time and showed an improved conflict resolution compared to Dempster's rule.
\citet{SEGM2020} adapt the \ac{BBA} obtained from \ac{lidar} and \glspl{stereocam} by adding a similarity factor and apply Dempster's rule to the adapted \acp{BBA}.
Compared to applying Dempster's rule to the original \acp{BBA}, they obtain a reduced entropy and a higher specificity.
\citet{Ullah2021} proposed a new uncertainty measure based on Deng's entropy and combine their entropy measures using Dempster's rule.
By doing so, they could increase the accuracy of the fusion results compared to Dempster's rule and using Deng's original entropy.

The critical cases when fusing heterogeneous sensor data in top-view grid maps occur if the \acp{BBA} obtained from the individual sensors are highly conflicting similar to the example demonstrated in \Cref{tab:dempster_example,tab:conjunctive_example}.
We propose the inclusion of information on the credibility of the individual sensors in order to resolve those conflicts.
This is not covered in the above-mentioned publications.
\section{Evidential Information Fusion}

Evidence Theory was formally introduced by Shafer in~\cite{Shafer1976} and provides a framework to model uncertainty and combine evidence degrees from independent sources.
Let $\Omega$ be a set consisting of mutually excluding hypotheses of interest called \ac{FOD} and $2^\Omega$ its power set containing all subsets $A\subseteq\Omega$.
The mapping
\begin{equation}
    \fbba \colon 2^\Omega \rightarrow \left[ 0, 1 \right]~, \quad \bba{\emptyset} = 0, \quad \sum_{A \in 2^\Omega} \bba{A} = 1
\end{equation}
is called \ac{BBA} and assigns a degree of evidence to all possible combinations of hypotheses.
In order to derive a probability measure from a \ac{BBA}, Smets proposed the pignistic transformation
\begin{equation}
    \pr{A} = \sum\limits_{B\subseteq\Omega}\frac{\vert A\cap B\vert}{\vert B\vert}\bba{B}
    \label{eq:pignistic}
\end{equation}
in \cite{Smets1990}.

Analogously to Shannon's entropy measure for probabilistic random variables, it has been ongoing research to define similar measures in the evidence theoretical context.
Deng's entropy
\begin{equation} \label{eq:deng_entropy}
    \denge{\fbba} = -\sum\limits_{A\subseteq\Omega} \bba{A} \log_2\left(\frac{\bba{A}}{2^{\vert A\vert} - 1}\right),
\end{equation}
presented in \cite{Deng2016} can be written as the sum of the nonspecificity
\begin{equation} \label{eq:deng_nonspecificity}
    \dengns{\fbba} = \sum\limits_{A\subseteq\Omega} \bba{A} \log_2\left(2^{\vert A\vert} - 1\right)
\end{equation}
and the discord
\begin{equation} \label{eq:deng_discord}
    \dengd{\fbba} = -\sum\limits_{A\subseteq\Omega} \bba{A} \log_2\left(\bba{A}\right).
\end{equation}
The nonspecificity $\dengns{\fbba}$ is a measure for the ignorance contained in the \ac{BBA} and increases with \ac{BBA} masses assigned to non singleton hypotheses.
The discord $\dengd{\fbba}$ is a measure for the indecision between several focal elements.
The upper and lower bound of Deng's entropy are the \ac{BBA} $\fbba_1$ assigning all the evidence mass $\fbba_1(\omega) = 1$ to one singleton hypothesis $\omega\in\Omega$ and the distribution of total ignorance $\fbba_2(\Omega) = 1$:
\begin{equation}
    \denge{\fbba_1} = 0 \leq \dengns{\fbba} \leq \log_2\left(2^{\vert\Omega\vert} - 1\right) = \denge{\fbba_2} < \vert\Omega\vert.
\end{equation}

In order to combine \acp{BBA} from independent sources, Dempster's rule of combination was introduced in \cite{Dempster1967}.
Here, two \acp{BBA} $\fbba_1$ and $\fbba_2$ are combined as
\begin{equation}
    (\fbba_1 \oplus \fbba_2)(A) = \frac{1}{1 - K} \sum\limits_{X\cap Y = A} \fbba_1(X)\,\fbba_2(Y), \label{eq:dempster_comb}
\end{equation}
where
\begin{equation}
    K = \sum\limits_{X\cap Y = \emptyset} \fbba_1(X)\,\fbba_2(Y)
\end{equation}
denotes the accumulated conflict mass.
The normalization constant $\frac{1}{1 - K}$ distributes the conflicts equally to all focal elements.
Dempster's rule satisfies desired properties as commutativity and associativity.
Although Dempster's rule is frequently used in literature, it has also been criticized.
Zadeh showed in \cite{Zadeh1979} that combining highly conflicting \acp{BBA} with Dempster's rule leads to counterintuitive results.
This effect is also known as Zadeh's paradox.
\Cref{tab:dempster_example} shows such an example.
Although both $\fbba_1$ and $\fbba_2$ hold high evidences masses against hypothesis $B$, Dempster's rule yields $(\fbba_1 \oplus \fbba_2)(B) = 1$ ignoring the conflict mass $K = 0.99$.
\begin{table}[ht]
    \centering
    \caption{Two \acp{BBA} $\fbba_1$ and $\fbba_2$ on $\Omega = \{A, B, C\}$ combined with Dempster's rule (\Cref{eq:dempster_comb}).}
    \begin{tabular}{c|c|c|c|c}
                                 & $A$   & $B$   & $C$   & $\Omega$ \\ \hline
        $\fbba_1$                & $0.9$ & $0.1$ & $0$   & $0$      \\ \hline
        $\fbba_2$                & $0$   & $0.1$ & $0.9$ & $0$      \\ \hline
        $\fbba_1 \oplus \fbba_2$ & $0$   & $1$   & $0$   & $0$      \\
    \end{tabular}
    \label{tab:dempster_example}
\end{table}

Different combination rules have been proposed aiming at resolving this counterintuitivity that all address different ways of dealing with conflicts.
\citet{Yager1987} defined the conjunctive rule of combination given as
\begin{equation}
    (\fbba_1 \ocap \fbba_2)(A) = \sum\limits_{X\cap Y = A} \fbba_1(X)\,\fbba_2(Y),
    \label{eq:conjunctive_comb}
\end{equation}
for $A \neq \Omega$, and
\begin{equation}
    (\fbba_1 \ocap \fbba_2)(A) = \fbba_1(\Omega)\,\fbba_2(\Omega) + \sum\limits_{X\cap Y = \emptyset} \fbba_1(X)\,\fbba_2(Y)
\end{equation}
for $A = \Omega$.
Note that it merely drops the normalization constant and assigns the conflict mass $K$ to $\Omega$ compared to Dempster's rule in \Cref{eq:dempster_comb}.
\begin{table}[ht]
    \centering
    \caption{Two \acp{BBA} $\fbba_1$ and $\fbba_2$ on $\Omega = \{A, B, C\}$ combined with the conjunctive rule (\Cref{eq:conjunctive_comb}).}
    \begin{tabular}{c|c|c|c|c}
                                & $A$   & $B$    & $C$   & $\Omega$ \\ \hline
        $\fbba_1$               & $0.9$ & $0.1$  & $0$   & $0$      \\ \hline
        $\fbba_2$               & $0$   & $0.1$  & $0.9$ & $0$      \\ \hline
        $\fbba_1 \ocap \fbba_2$ & $0$   & $0.01$ & $0$   & $0.99$   \\
    \end{tabular}
    \label{tab:conjunctive_example}
\end{table}
\Cref{tab:conjunctive_example} shows the results when applying the conjunctive rule to the example introduced in \Cref{tab:dempster_example}.
Compared to Dempster's rule the conjunctive rule assigns the conflict mass $K$ to $(\fbba_1 \ocap \fbba_2)(\Omega)$ indicating a high degree of uncertainty.
Although the conjunctive rule gives a more intuitive result in this example it discards a significant amount of information by assigning the whole conflict mass to $\Omega$.
Other examples for modified combination rules are Duboi's and Prade's rule presented in \cite{Dubois1988} and the \ac{PCR} rules introduced in \cite{Smarandache2006,Smarandache2006b}.
All of them are based on Yager's rule (\Cref{eq:conjunctive_comb}) and assign conflict masses in different ways.

\citet{Yang2013} propose another approach to dealing with conflicts when combining \acp{BBA}.
Given the \ac{FOD} $\Omega$, they consider sources of evidence $\{e_i, i=1,\dots,n\}$ with weight $0\leq w_i\leq 1$ and reliability $0\leq r_i\leq 1$ each providing a \ac{BBA} $\fbba_i$.
The weight models the relative importance of the source of evidence, whereas the reliability models the information quality.
The \ac{BBA} $\fbba_i$ of the source of evidence $e_i$ is modified based on the weight $w_i$ and the reliability $r_i$ as
\begin{equation}
    \tilde{\fbba}_i(A) = \begin{cases}
        0, & \text{ if } A = \emptyset, \\
        \frac{1}{1+w_i-r_i}\,\fbba_i(A), & \text{ if } A \in \powerset{\Omega}\setminus\emptyset.
    \end{cases}
    \label{eq:wbd}
\end{equation}
Two independent sources of evidence $e_1$ and $e_2$ with reliabilities $0 \leq r_1, r_2 \leq 1$ and modified \acp{BBA} $\tilde{\fbba}_1$ and $\tilde{\fbba}_2$ are then combined as
\begin{equation}
    \fbba_{1, 2}(A) = \begin{cases}
        0, & \text{ if } A = \emptyset, \\
        \frac{\tilde{\fbba}_{1,2}(A)}{\sum_{B\in \powerset{\Omega}} \tilde{\fbba}_{1,2}(B)}, & \text{ if } A \in \powerset{\Omega}\setminus\emptyset,
    \end{cases}
    \label{eq:wer_comb}
\end{equation}
where
\begin{equation}
    \begin{split}
    \tilde{\fbba}_{1,2}(A) =    & \,(1 - r_2)\,\tilde{\fbba}_1(A) + (1 - r_1)\,\tilde{\fbba}_2(A) \\
                                & + \sum\limits_{B\cap C=A}\tilde{\fbba}_1(B)\,\tilde{\fbba}_2(C).
    \end{split}
\end{equation}
Note that for $w_1=w_2=r_1=r_2=1$, \Cref{eq:wer_comb} reduces to Dempster's rule.
In the remainder of this paper, \Cref{eq:wer_comb} will be referred to as \acf{ER} rule.
The two reliability parameters $r_1,r_2$ influence the combination results significantly.
\begin{table}[ht]
    \centering
    \caption{Two \acp{BBA} $\fbba_1$ and $\fbba_2$ on $\Omega = \{A, B, C\}$ combined with the \ac{ER} rule (\Cref{eq:wer_comb}).}
    \begin{tabular}{c|c|c|c|c}
                                    & $A$   & $B$   & $C$       & $\Omega$  \\ \hline
        $\fbba_1$ with $r_1=0.7$    & $0.9$ & $0.1$ & $0$       & $0$       \\ \hline
        $\fbba_2$ with $r_2=0.3$    & $0$   & $0.1$ & $0.9$     & $0$       \\ \hline
        $\fbba_1 \ocap \fbba_2$     & $0.67$& $0.11$& $0.22$    & $0$       \\
    \end{tabular}
    \label{tab:wer_example}
\end{table}
\Cref{tab:wer_example} shows the results when combining the two \acp{BBA} from \Cref{tab:dempster_example,tab:conjunctive_example} with the \ac{ER} combination rule where the reliabilities were set to $r_1=0.7$ and $r_2=0.3$ and $w_1=w_2=1$.
Due to the higher reliability assigned to $\fbba_1$, the conflict between $A$ and $C$ is mostly assigned to $A$.

In summary, evidential reasoning with the \ac{ER} combination rule (\Cref{eq:wer_comb}) provides a mathematical framework for dealing with differently credible sources of evidence without losing the mathematical properties of Dempster's original combination rule.

\section{Methodology}

To the best of the author's knowledge the \ac{ER} combination rule has not yet been applied to the fusion of evidential grid maps.
Subsequently, we first present our evidential grid map model and explain how the sensor grid maps are combined using \ac{ER}.

\subsection{Evidential Grid Map Model}

In this work, we consider the semantic classes ``car'', ``two-wheeler'', ``pedestrian'', ``other mobile obstacles'' or ``immobile obstacles''.
In particular, the occupancy \ac{FOD}
\begin{equation}
    \frameo \defeq \{ c, cy, p, m, nm, f, v \}
    \label{eq:fod_o}
\end{equation}
consists of the hypotheses displayed in \Cref{tab:hypotheses_obstacle}.
\begin{table}[!ht]
	\caption{The \ac{FOD} $\frameo$ used in the evidential grid map representation. 
            The individual semantic hypotheses refine the classical occupancy hypothesis.}
    \centering
	\begin{tabular}{lcc}
		Semantic class 	& Set \\
		\hline
		Occupied by\ldots      			    &       	                   \\
		\ldots car 				            & $\{c\}$	         \\
		\ldots two-wheeler 			        & $\{cy\}$     \\
		\ldots pedestrian 		            & $\{p\}$    \\
		\ldots other mobile object 	        & $\{om\}$      \\
		\ldots immobile object              & $\{nm\}$   \\
        \ldots unknown object type          & $\{c, cy, p, om, nm\}$   \\
		\hline
		Free 			                    & $\{f\}$ 
	\end{tabular}
	\label{tab:hypotheses_obstacle}
\end{table}
The \ac{BBA} $\mathrm{m}$ on $2^{\frameo}$ is then represented by the multi-layer grid map
\begin{equation}
    \gm \colon \grid \times 2^{\frameo}\rightarrow [0, 1].
\end{equation}
Here, the \ac{2D} Cartesian grid $\grid = \mathcal{P}_1\times \mathcal{P}_2$ forms a partition of the rectangular region of interest $\rect = I_1\times I_2 \subset \real^2$, where
\begin{align*}
    \mathcal{P}_i & = \{I_{i, k},\, k\in \{0,\dots,s_i-1\}\},                      \\
    I_{i, k}      & =[o_i + k\,\delta_i, o_i + (k+1)\,\delta_i),\quad i\in\{1, 2\},
\end{align*}
with grid cell side length $\delta_i \in\real$, origin $o_i\in \real$ and size $s_i\in\mathbb{N}$.

\subsection{Evidential Reasoning with Reliability}

We apply the \ac{ER} combination rule presented by \citet{Yang2013} to \glspl{sensorgm} and model the reliability $r_i$ of sensor sources to improve conflict resolution.
The importance weights $w_i$ are set to one modeling all sources to be equally important.
Given two \glspl{sensorgm} $\gm_1$ and $\gm_2$ calculated using measurements from two independent sensors $s_1$ and $s_2$, the combination $\gm_{1,2}$ is computed.
The two sensors $s_1$ and $s_2$ are interpreted as sources of evidence with reliabilities $0\leq r_1, r_2\leq 1$.
In a fixed grid cell $\cell\in\grid$, the \acp{BBA} $\gm_1(\cell, \,\cdot\,)$ and $\gm_2(\cell, \,\cdot\,)$ are then combined to the \ac{BBA} $\gm_{1,2}(\cell, \,\cdot\,)$ using the \ac{ER} rule.

The reliability of a source of evidence should be chosen carefully.
Let $\fbba_1$ and $\fbba_2$ be two \acp{BBA} to be combined.
We propose to model the reliabilities $r_i$ as a function of the conflict mass
\begin{equation}
    K = \sum\limits_{X\cap Y = \emptyset} \fbba_1(X)\,\fbba_2(Y)
\end{equation}
and the credibility coefficient $b_i$ as 
\begin{equation}
    r_i = \func{f}_r(K, b_i) = 1 - (1 - b_i)\,K.
\end{equation}
The coefficient $0\leq b_i\leq 1$ models the credibility of a source of evidence in light of a conflict.
If $K=0$, then $r_1=r_2=1$ and the \ac{ER} rule reduces to Dempster's rule.
The higher the conflict mass $K$, the more unintuitive the combination result with Dempster's rule becomes as shown in \Cref{tab:dempster_example} and the combination rule is adapted.
If $K=1$, we have $r_i = b_i$ and the credibility coefficient fully serves as reliability value.

In this work we apply the presented general framework to the combination of grid maps estimated based on one \ac{lidar} scanner and one \gls{stereocam} each representing the \ac{BBA} on the \ac{FOD} $\frameo$. 

\subsection{Parameter Estimation}

When combining \ac{lidar} and \gls{stereocam} \glspl{sensorgm} with the \ac{ER} rule, the sensor credibility coefficients $b_l$ for the \ac{lidar} and $b_s$ for the \gls{stereocam} need to be specified.
In this work, we propose a data-driven approach assigning the values $b_l, b_s$ resulting in the best fusion performance in terms of a quantitative evaluation.
We use the \ac{3D} semantic bounding primitives and the semantic point cloud in the \kittigeiger{} dataset to generate a reference semantic evidential grid map $g_\gt$.
\kittigeiger{} accumulates \ac{lidar} measurement chunks containing over 300 frames.
Each frame includes \ac{lidar} reflections with a distance of at most 30\si{\metre}.
The accumulated point clouds were semantically annotated and used to fit bounding primitives representing object instances and infrastructural entities such as buildings, poles and streets.
The labeled bounding primitives and the accumulated point clouds are projected into the reference grid map $g_\gt$ representing the reference \ac{BBA} per grid cell.
As a quality measure, we define the \ac{eIoU}
\begin{equation}
    \eiou_\omega = \frac{\etp_\omega}{\etp_\omega+\efp_\omega+\efn_\omega}
    \label{eq:eiou}
\end{equation}
based on the evidential true positive rate
\begin{equation}
    \etp_\omega = \sum_{\cell\in\mathcal{G}_{xy}}\sum_{\phi \cap \omega = \phi} g_\gt(\cell, \phi)\,g_\mathcal{M}\left ( \cell, \omega \right ),
\end{equation}
the false positive rate
\begin{equation}
    \efp_\omega = \sum_{\cell\in\mathcal{G}_{xy}}\sum_{\phi \cap \omega = \emptyset} g_\gt(\cell, \phi)\,g_\mathcal{M}\left ( \cell, \omega \right ),
\end{equation}
and false negative rate
\begin{equation}
    \efn_\omega = \sum_{\cell\in\mathcal{G}_{xy}}\sum_{\phi \cap \omega = \emptyset} g_\gt(\cell, \omega)\,g_\mathcal{M}\left ( \cell, \phi \right ).
\end{equation}
Note that the \ac{eIoU} reduces to the classical \ac{IoU} used for pixel-wise semantic labeling if the \ac{BBA} is binary, i.e. $\bba{\omega} = 1$ for one $\omega\in2^\Omega$.

Now, the performance of the information fusion is quantified by the \ac{eIoU} (\Cref{eq:eiou}) for the hypotheses \emph{occupied by unknown object type}.
More specificity, we apply the \ac{ER} rule to \ac{lidar} and \gls{stereocam} \glspl{sensorgm} based on measurements in the \gls{kitti360} dataset \cite{Liao2021ARXIV} for credibility values 
\begin{equation*}
    \{(b_l,b_s)\,\vert\, b_l, b_s \in 0.1\cdot\mathbb{N}_{\leq 10}\}.
\end{equation*}
In this context, all the occupancy evidence is assigned to the hypothesis \emph{occupied by unknown object type} as the individual semantic hypotheses are not considered in the analysis.
\begin{figure}[!t]
    \fontsize{8pt}{8pt}\selectfont
    \providecommand{\figheight}{2in}%
    \providecommand{\figwidth}{3in}%
    \input{figures/credibility_ious.tex}
    \caption{The \acp{eIoU} of the hypotheses \emph{occupied by unknown object type} for different combinations of credibility coefficients $(b_l,b_s)$.
            The highest \ac{eIoU} is achieved for $b_l = 1$ and $b_s = 0$.}
    \label{fig:credibility_ious}
\end{figure}
The results are shown in the table in \Cref{fig:credibility_ious}.
Each entry contains the \ac{eIoU} averaged over all cell states in 50 frames of the \gls{kitti360} dataset.
It can be seen that the \ac{eIoU} increases with increasing \ac{lidar} credibility $b_l$, independent of the \gls{stereocam} credibility.
Furthermore, the \acp{eIoU} increases for decreasing \gls{stereocam} credibilities.
The highest \ac{eIoU} is measured for the \ac{lidar} credibility $b_l=1$ and the \gls{stereocam} credibility $b_s=0$.
For this combination of credibility values, the \ac{eIoU} is $0.41$ percentage points higher compared to applying Dempster's rule and $2.9$ percentage points higher than with $b_l = b_s = 0$.
This difference is quite significant considering that the sensor estimates are not conflicting in the majority of the grid cells and both rules coincide in those cases.
Recall that $b_s=0$ does not mean that the \ac{BBA} estimated with the \gls{stereocam} is not regarded at all in the combination.
It merely means that in cases where the \gls{stereocam} provides a measurement that disagrees with the \ac{lidar} measurement, the \ac{lidar} measurement shall be considered more reliable.
The \ac{eIoU}-based analysis shows how close the fusion result is to the reference grid map based on the semantic labels and bounding box primitives in the \gls{kitti360} dataset.
As those labels were annotated using the \ac{lidar} measurements this result does not really come as a surprise.
However, this demonstrates that the sensor fusion results can be tuned as desired by adapting the credibility coefficients in the \ac{ER} rule.
As we consider the annotations in the \gls{kitti360} dataset to be accurate in this work, we therefore set $b_l = 1$ and $b_s = 0$ for the remainder of this paper.
This is sensible as in fact, \ac{lidar} scanners provide more accurate depth estimates compared to the disparity-based depth estimation with \glspl{stereocam}.

\section{Experiments}

We evaluate the fusion results qualitatively and quantitatively based on \lidarkitti{} \ac{lidar} and \gls{stereocam} measurements in the \gls{kitti360} dataset presented in \cite{Liao2021ARXIV}.

\subsection{Qualitative Results}

We show qualitative results for the fusion of \glspl{sensorgm} from \lidarkitti{} \ac{lidar} scans without semantic estimates with \glspl{sensorgm} from disparity maps from stereo images with semantic estimates.

\begin{figure}[!t]
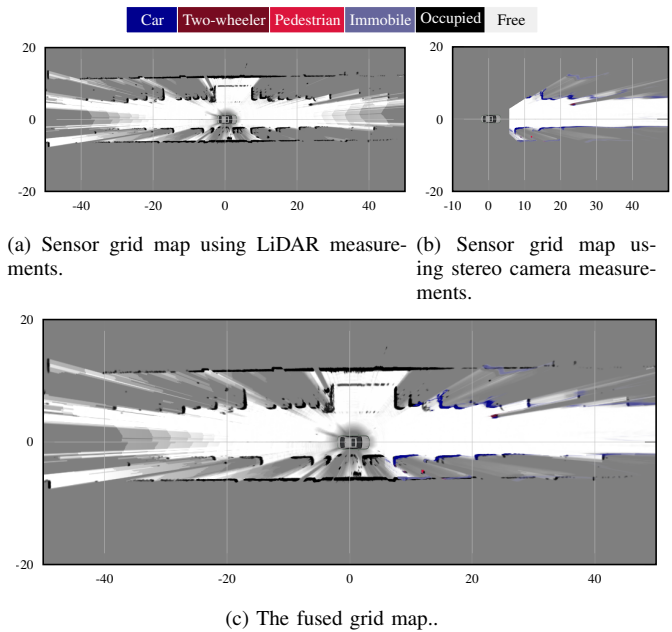

    \centering
    \colorbarev
    \subcaptionbox{\Gls{sensorgm} using \ac{lidar} measurements.\label{fig:fused_bba_vis_sem_lidar}}%
    [.6\linewidth]{\widegrid{file=lidar_bba_vis_kitti_odom_00_3618,width=0.9}}
    \subcaptionbox{\Gls{sensorgm} using \gls{stereocam} measurements.\label{fig:fused_bba_vis_sem_stereo}}%
    [.36\linewidth]{\widegridcam{file=stereo_bba_vis_sem_no_ground_rect_kitti_odom_00_3618,width=0.9}}
    \subcaptionbox{The \gls{fusedgm}..\label{fig:fused_bba_vis_sem}}%
    [\linewidth]{\widegrid{file=fused_bba_vis_sem_no_ground_kitti_odom_00_3618,width=0.92}}
    \caption{Resulting evidential grid map for the fusion of \ac{lidar} measurements without and \gls{stereocam} measurements with semantic estimates using the \ac{ER} rule.}
    \label{fig:fused_bba}
\end{figure}
\Cref{fig:fused_bba} shows the grid maps for one frame.
In the sensor grid map based on \ac{lidar} measurements that is depicted in \Cref{fig:fused_bba_vis_sem_lidar}, \ac{BBA} estimates are only available for the hypotheses \emph{free} and \emph{occupied by unknown object type}.
The grid map based on \gls{stereocam} measurements depicted in \Cref{fig:fused_bba_vis_sem_stereo} contains \ac{BBA} estimates for the individual semantic hypotheses based on the semantic labeling provided by the neural network.
Both \glspl{sensorgm} are estimated using the method described in \cite{sensorgm} where the \gls{stereocam} grid map is based on stereo disparities obtained from the guided aggregation net for stereo matching presented by \citet{ganet} and the pixel-wise semantic labels are estimated using the network presented by \citet{zhu2019improving}.
It can be seen that the spatial uncertainty is higher in the \gls{stereocam} grid map indicated by a more blurry occupancy pattern.
The result of combining the two \glspl{sensorgm} with the \ac{ER} rule is shown in \Cref{fig:fused_bba_vis_sem}.
The semantic estimates provided by the \gls{stereocam} is successfully included in the occupancy pattern obtained from the \ac{lidar} scanner.

\begin{figure}[!t]
    \fontsize{8pt}{8pt}\selectfont
    \includesvg[width=\linewidth]{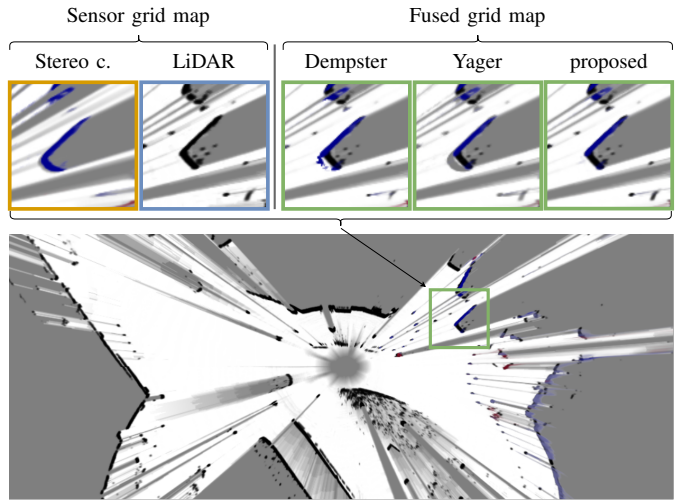}
    \caption{Results for fusing \ac{lidar} grid maps with \gls{stereocam} grid maps using different combination rules.
            A subarea is highlighted where a vehicle is located.
            The measurement conflict between the \acp{BBA} provided by the two sensors is resolved differently.
            Only the proposed method resolves the conflict correctly in most of the grid cells.}
    \label{fig:fused_bba_comp}
\end{figure}
In order to demonstrate the effect of using different evidential combination rules in case of conflicting sensor \acp{BBA}, the same sensor grid maps are combined using Dempster's rule (\Cref{eq:dempster_comb}), Yager's rule (\Cref{eq:conjunctive_comb}) and the \ac{ER} rule (\Cref{eq:wer_comb}) in \Cref{fig:fused_bba_comp}.
In this example, the distance of an observed car was underestimated by the stereo pipeline leading to an occupied-free conflict in front of the car.
With Dempster's rule, conflicting \acp{BBA} are distributed equally over all focal elements.
Hence, large parts of the conflict mass are assigned to the hypothesis car.
Yager's rule assigns all conflict masses to $\Omega$, thus leading to a low \ac{BBA} for both hypotheses free and car.
The \ac{ER} rule on the other hand is able to correctly assign large parts of the conflict masses to the hypothesis free due to the lower credibility coefficient $r_s$ assigned to the \gls{stereocam}.

\subsection{Quantitative Evaluation}

We evaluate the accuracy of the \ac{BBA} by calculating the \ac{eIoU} in \num{3000} frames of the \gls{kitti360} dataset for the fused grid maps and comparing it to the results obtained for the sensor measurement grid maps.
The hypotheses \emph{occupied by immobile object} and \emph{occupied by other mobile object} are not considered here as the reference \ac{BBA} for the former was found to be not credible, and the latter is barely observed in the evaluation dataset.

\begin{figure}[!t]
    \centering
    \fontsize{8pt}{8pt}\selectfont
    \providecommand{\figheight}{1.35in}%
    \providecommand{\figwidth}{0.87\linewidth}%
    % This file was created with tikzplotlib v0.9.15.
\begin{tikzpicture}

\definecolor{color0}{rgb}{0.392156862745098,0.392156862745098,0.949019607843137}
\definecolor{color1}{rgb}{0.466666666666667,0.0431372549019608,0.125490196078431}
\definecolor{color2}{rgb}{0.862745098039216,0.0784313725490196,0.235294117647059}

\begin{axis}[
axis line style={white},
height=\figheight,
legend cell align={left},
legend columns=5,
legend style={fill opacity=0.8, draw opacity=1, text opacity=1, at={(0.5,1)}, anchor=south, draw=none},
minor xtick={},
minor ytick={},
tick align=outside,
tick pos=left,
width=\figwidth,
x grid style={white!79.6078431372549!black},
xmajorgrids,
xmin=-0.2, xmax=4.2,
xtick style={color=black},
xtick={0,1,2,3,4},
xticklabels={car,two-wheeler,pedestrian,occupied,free},
y grid style={white!79.6078431372549!black},
ymajorgrids,
ymin=0, ymax=103.11,
ytick style={color=black},
ytick={0,20,40,60,80,100,120}
]
\draw[draw=none,fill=color0,very thin,postaction={pattern=north east lines}] (axis cs:-0.3915,0) rectangle (axis cs:-0.1485,0);
\addlegendimage{area legend,draw=none,fill=color0,very thin,postaction={pattern=north east lines}}
\addlegendentry{\acs{lidar}}

\draw[draw=none,fill=color1,very thin,postaction={pattern=north east lines}] (axis cs:0.6085,0) rectangle (axis cs:0.8515,0);
\draw[draw=none,fill=color2,very thin,postaction={pattern=north east lines}] (axis cs:1.6085,0) rectangle (axis cs:1.8515,0);
\draw[draw=none,fill=white!19.6078431372549!black,very thin,postaction={pattern=north east lines}] (axis cs:2.6085,0) rectangle (axis cs:2.8515,91.6);
\draw[draw=none,fill=white!90.1960784313726!black,very thin,postaction={pattern=north east lines}] (axis cs:3.6085,0) rectangle (axis cs:3.8515,98.2);
\draw[draw=none,fill=color0,very thin,postaction={pattern=north west lines}] (axis cs:-0.1215,0) rectangle (axis cs:0.1215,43.4);
\addlegendimage{area legend,draw=none,fill=color0,very thin,postaction={pattern=north west lines}}
\addlegendentry{\gls{stereocam}}

\draw[draw=none,fill=color1,very thin,postaction={pattern=north west lines}] (axis cs:0.8785,0) rectangle (axis cs:1.1215,35.4);
\draw[draw=none,fill=color2,very thin,postaction={pattern=north west lines}] (axis cs:1.8785,0) rectangle (axis cs:2.1215,14.1);
\draw[draw=none,fill=white!19.6078431372549!black,very thin,postaction={pattern=north west lines}] (axis cs:2.8785,0) rectangle (axis cs:3.1215,1.6);
\draw[draw=none,fill=white!90.1960784313726!black,very thin,postaction={pattern=north west lines}] (axis cs:3.8785,0) rectangle (axis cs:4.1215,97.3);
\draw[draw=none,fill=color0,very thin,postaction={pattern=horizontal lines}] (axis cs:0.1485,0) rectangle (axis cs:0.3915,46.1);
\addlegendimage{area legend,draw=none,fill=color0,very thin,postaction={pattern=horizontal lines}}
\addlegendentry{fused}

\draw[draw=none,fill=color1,very thin,postaction={pattern=horizontal lines}] (axis cs:1.1485,0) rectangle (axis cs:1.3915,38.5);
\draw[draw=none,fill=color2,very thin,postaction={pattern=horizontal lines}] (axis cs:2.1485,0) rectangle (axis cs:2.3915,17.1);
\draw[draw=none,fill=white!19.6078431372549!black,very thin,postaction={pattern=horizontal lines}] (axis cs:3.1485,0) rectangle (axis cs:3.3915,91.4);
\draw[draw=none,fill=white!90.1960784313726!black,very thin,postaction={pattern=horizontal lines}] (axis cs:4.1485,0) rectangle (axis cs:4.3915,97.5);
\draw (axis cs:-0.27,0) ++(0pt,0pt) node[
  scale=0.7,
  anchor=south,
  text=black,
  rotate=0.0
]{0};
\draw (axis cs:0.73,0) ++(0pt,0pt) node[
  scale=0.7,
  anchor=south,
  text=black,
  rotate=0.0
]{0};
\draw (axis cs:1.73,0) ++(0pt,0pt) node[
  scale=0.7,
  anchor=south,
  text=black,
  rotate=0.0
]{0};
\draw (axis cs:2.73,91.6) ++(0pt,0pt) node[
  scale=0.7,
  anchor=south,
  text=black,
  rotate=0.0
]{91.6};
\draw (axis cs:3.73,98.2) ++(0pt,0pt) node[
  scale=0.7,
  anchor=south,
  text=black,
  rotate=0.0
]{98.2};
\draw (axis cs:0,43.4) ++(0pt,0pt) node[
  scale=0.7,
  anchor=south,
  text=black,
  rotate=0.0
]{43.4};
\draw (axis cs:1,35.4) ++(0pt,0pt) node[
  scale=0.7,
  anchor=south,
  text=black,
  rotate=0.0
]{35.4};
\draw (axis cs:2,14.1) ++(0pt,0pt) node[
  scale=0.7,
  anchor=south,
  text=black,
  rotate=0.0
]{14.1};
\draw (axis cs:3,1.6) ++(0pt,0pt) node[
  scale=0.7,
  anchor=south,
  text=black,
  rotate=0.0
]{1.6};
\draw (axis cs:4,97.3) ++(0pt,0pt) node[
  scale=0.7,
  anchor=south,
  text=black,
  rotate=0.0
]{97.3};
\draw (axis cs:0.27,46.1) ++(0pt,0pt) node[
  scale=0.7,
  anchor=south,
  text=black,
  rotate=0.0
]{46.1};
\draw (axis cs:1.27,38.5) ++(0pt,0pt) node[
  scale=0.7,
  anchor=south,
  text=black,
  rotate=0.0
]{38.5};
\draw (axis cs:2.27,17.1) ++(0pt,0pt) node[
  scale=0.7,
  anchor=south,
  text=black,
  rotate=0.0
]{17.1};
\draw (axis cs:3.27,91.4) ++(0pt,0pt) node[
  scale=0.7,
  anchor=south,
  text=black,
  rotate=0.0
]{91.4};
\draw (axis cs:4.27,97.5) ++(0pt,0pt) node[
  scale=0.7,
  anchor=south,
  text=black,
  rotate=0.0
]{97.5};
\end{axis}

\end{tikzpicture}    
    \caption{The \acp{eIoU} in \% for five hypotheses in the \ac{FOD} $\frameo$. 
            For each hypothesis, the bar plots show the \acp{eIoU} measured in the grid map based on \ac{lidar} measurements, \gls{stereocam} measurements and for the fused grid map.
            The \ac{BBA} from the \gls{stereocam} contains estimates for the semantic occupancy hypotheses whereas the \ac{lidar}-based \ac{BBA} only contains evidence for the hypotheses \emph{occupied by unknown object type} and \emph{free}.}
    \label{fig:fusion_eiou_objects}
\end{figure}
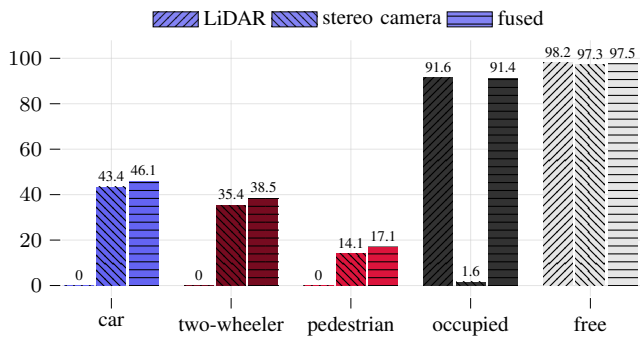
\Cref{fig:fusion_eiou_objects} shows the results for the \ac{FOD} $\frameo$ in the described sensor setup.
Because no semantic estimates are included in the \ac{lidar}-based estimation chain, the \ac{eIoU} values are zero for all semantic occupancy hypotheses.
As expected, it can be seen that the \ac{lidar} scanner leads to a more accurate \ac{BBA} estimation than the \gls{stereocam} for the hypotheses estimated by both sensors.
Furthermore, the metrics show that the accuracy for the singleton hypotheses \emph{car}, \emph{two-wheeler} and \emph{pedestrian} can be significantly improved by fusing the \gls{occsem} estimation obtained from the \gls{stereocam} with the occupancy information obtained from the \ac{lidar}.
Although no further semantic estimates are added in the fusion process the \ac{eIoU} can be improved by 6.2\% for \emph{cars}, by 8.8\% for \emph{two-wheelers} and by 21.3\% for \emph{pedestrians}.
This is due to the improved conflict resolution when applying the \ac{ER} combination rule as verified in \Cref{fig:fused_bba_comp}.

In addition to the comparison with reference \ac{BBA} maps, the uncertainty incorporated in the estimated \ac{BBA} is evaluated based on Deng's entropy measures defined in \Cref{eq:deng_entropy,eq:deng_nonspecificity,eq:deng_discord}.
\begin{figure}[t]
    \centering
    \fontsize{8pt}{8pt}\selectfont
    \providecommand{\figheight}{0.85in}%
    \providecommand{\figwidth}{0.69\linewidth}%
    \subcaptionbox{360° view.\label{fig:fusion_sem_obj_uncertainty_whole}}%
    [.498\linewidth]{% This file was created with tikzplotlib v0.9.15.
\begin{tikzpicture}

\definecolor{color0}{rgb}{0.392156862745098,0.584313725490196,0.929411764705882}
\definecolor{color1}{rgb}{0.941176470588235,0.501960784313725,0.501960784313725}

\begin{axis}[
axis line style={white},
height=\figheight,
legend cell align={left},
legend columns=2,
legend style={fill opacity=0.8, draw opacity=1, text opacity=1, at={(0.5,1)}, anchor=south, draw=none},
minor xtick={},
minor ytick={},
tick align=outside,
tick pos=left,
width=\figwidth,
x grid style={white!79.6078431372549!black},
xmajorgrids,
xmin=2, xmax=5.6,
xtick style={color=black},
xtick={2,3,4,5,6},
y dir=reverse,
y grid style={white!79.6078431372549!black},
ymajorgrids,
ymin=-0.5, ymax=2.3,
ytick style={color=black},
ytick={0,1,2},
yticklabels={\acs{lidar},stereo c.,fused}
]
\draw[draw=none,fill=color0,very thin] (axis cs:0,-0.25) rectangle (axis cs:3.37,0.25);
\addlegendimage{area legend,draw=none,fill=color0,very thin}
\addlegendentry{$\dengns{\fbba}$}

\draw[draw=none,fill=color0,very thin] (axis cs:0,0.75) rectangle (axis cs:5.09,1.25);
\draw[draw=none,fill=color0,very thin] (axis cs:0,1.75) rectangle (axis cs:3.22,2.25);
\draw[draw=none,fill=color1,very thin] (axis cs:3.37,-0.25) rectangle (axis cs:3.64,0.25);
\addlegendimage{area legend,draw=none,fill=color1,very thin}
\addlegendentry{$\dengd{\fbba}$}

\draw[draw=none,fill=color1,very thin] (axis cs:5.09,0.75) rectangle (axis cs:5.16,1.25);
\draw[draw=none,fill=color1,very thin] (axis cs:3.22,1.75) rectangle (axis cs:3.49,2.25);
\draw (axis cs:2.685,-0.375) node[
  scale=0.7,
  text=black,
  rotate=0.0
]{3.37};
\draw (axis cs:3.545,0.625) node[
  scale=0.7,
  text=black,
  rotate=0.0
]{5.09};
\draw (axis cs:2.61,1.625) node[
  scale=0.7,
  text=black,
  rotate=0.0
]{3.22};
\draw (axis cs:3.505,-0.375) node[
  scale=0.7,
  text=black,
  rotate=0.0
]{0.27};
\draw (axis cs:5.125,0.625) node[
  scale=0.7,
  text=black,
  rotate=0.0
]{0.07};
\draw (axis cs:3.355,1.625) node[
  scale=0.7,
  text=black,
  rotate=0.0
]{0.27};
\draw (axis cs:3.64,0) ++(0pt,0pt) node[
  scale=0.7,
  anchor=west,
  text=black,
  rotate=0.0
]{3.64};
\draw (axis cs:5.16,1) ++(0pt,0pt) node[
  scale=0.7,
  anchor=west,
  text=black,
  rotate=0.0
]{5.16};
\draw (axis cs:3.49,2) ++(0pt,0pt) node[
  scale=0.7,
  anchor=west,
  text=black,
  rotate=0.0
]{3.49};
\end{axis}

\end{tikzpicture}}\hfill
    \subcaptionbox{Camera view.\label{fig:fusion_sem_obj_uncertainty_cam_view}}%
    [.498\linewidth]{% This file was created with tikzplotlib v0.9.15.
\begin{tikzpicture}

\definecolor{color0}{rgb}{0.392156862745098,0.584313725490196,0.929411764705882}
\definecolor{color1}{rgb}{0.941176470588235,0.501960784313725,0.501960784313725}

\begin{axis}[
axis line style={white},
height=\figheight,
legend cell align={left},
legend columns=2,
legend style={fill opacity=0.8, draw opacity=1, text opacity=1, at={(0.5,1)}, anchor=south, draw=none},
minor xtick={},
minor ytick={},
tick align=outside,
tick pos=left,
width=\figwidth,
x grid style={white!79.6078431372549!black},
xmajorgrids,
xmin=2, xmax=3.1,
xtick style={color=black},
xtick={2,2.5,3,3.5},
y dir=reverse,
y grid style={white!79.6078431372549!black},
ymajorgrids,
ymin=-0.5, ymax=2.3,
ytick style={color=black},
ytick={0,1,2},
yticklabels={\acs{lidar},stereo c.,fused}
]
\draw[draw=none,fill=color0,very thin] (axis cs:0,-0.25) rectangle (axis cs:2.71,0.25);
\addlegendimage{area legend,draw=none,fill=color0,very thin}
\addlegendentry{$\dengns{\fbba}$}

\draw[draw=none,fill=color0,very thin] (axis cs:0,0.75) rectangle (axis cs:2.32,1.25);
\draw[draw=none,fill=color0,very thin] (axis cs:0,1.75) rectangle (axis cs:2.11,2.25);
\draw[draw=none,fill=color1,very thin] (axis cs:2.71,-0.25) rectangle (axis cs:2.95,0.25);
\addlegendimage{area legend,draw=none,fill=color1,very thin}
\addlegendentry{$\dengd{\fbba}$}

\draw[draw=none,fill=color1,very thin] (axis cs:2.32,0.75) rectangle (axis cs:2.59,1.25);
\draw[draw=none,fill=color1,very thin] (axis cs:2.11,1.75) rectangle (axis cs:2.36,2.25);
\draw (axis cs:2.355,-0.375) node[
  scale=0.7,
  text=black,
  rotate=0.0
]{2.71};
\draw (axis cs:2.16,0.625) node[
  scale=0.7,
  text=black,
  rotate=0.0
]{2.32};
\draw (axis cs:2.055,1.625) node[
  scale=0.7,
  text=black,
  rotate=0.0
]{2.11};
\draw (axis cs:2.83,-0.375) node[
  scale=0.7,
  text=black,
  rotate=0.0
]{0.24};
\draw (axis cs:2.455,0.625) node[
  scale=0.7,
  text=black,
  rotate=0.0
]{0.27};
\draw (axis cs:2.235,1.625) node[
  scale=0.7,
  text=black,
  rotate=0.0
]{0.25};
\draw (axis cs:2.95,0) ++(0pt,0pt) node[
  scale=0.7,
  anchor=west,
  text=black,
  rotate=0.0
]{2.95};
\draw (axis cs:2.59,1) ++(0pt,0pt) node[
  scale=0.7,
  anchor=west,
  text=black,
  rotate=0.0
]{2.59};
\draw (axis cs:2.36,2) ++(0pt,0pt) node[
  scale=0.7,
  anchor=west,
  text=black,
  rotate=0.0
]{2.36};
\end{axis}

\end{tikzpicture}}
    \caption{Nonspecificity $\dengns{\fbba}$ and discord $\dengd{\fbba}$ for \ac{lidar}, \gls{stereocam} and fused grid maps.}
    \label{fig:fusion_sem_obj_uncertainty}
\end{figure}
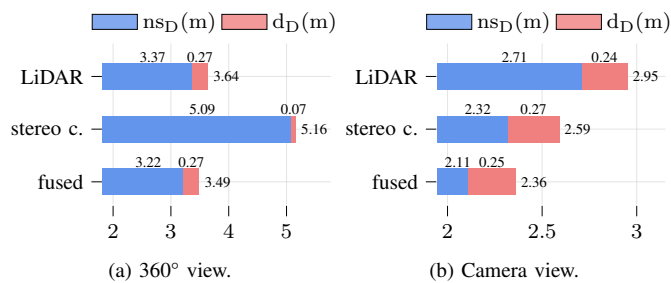
\Cref{fig:fusion_sem_obj_uncertainty} shows Deng's nonspecificity, discord and entropy averaged over the same \num{3000} frames of the \gls{kitti360} dataset for \ac{lidar}, \gls{stereocam} and fused measurements.
For each grid map, either all grid cells within a distance of 30 \si{\metre} to the ego vehicle (360° view) or the grid cells that are additionally withing the viewing are of the \gls{stereocam} (camera view), respectively, are taken into account.
In the 360° view shown in \Cref{fig:fusion_sem_obj_uncertainty_whole}, we observe that the average entropy is highest in the \gls{stereocam} grid maps.
This is expected as the \gls{stereocam} does not provide measurements outside the \gls{stereocam} view.
After applying the proposed sensor data fusion, the entropy is reduced by 4.1\% compared to the \ac{lidar} grid map.
The competitive part of the sensor data fusion is evaluated in the overlapping viewing areas of the sensors.
The corresponding entropy measures are plotted in \Cref{fig:fusion_sem_obj_uncertainty_cam_view}.
Here, the entropy in the \ac{lidar} grid map is higher than the entropy in the \gls{stereocam}.
Again the best result is obtained after applying the proposed sensor data fusion.
Due to a significant reduction of the nonspecificity, the entropy in the fused grid map is reduced by 8.9\% compared to the \gls{stereocam} grid map.
This demonstrates that the introduced fusion operator successfully aggregates information from different sources while keeping the discord at a constant level.

\section{Conclusion}

We proposed to apply \acf{ER} to sensor data fusion in top-view grid maps.
By resolving sensor conflicts adaptively based on reliability coefficients, we achieved accurate and robust fusion results even for highly conflicting sensor measurements.
The reliabilities for one \ac{lidar} scanner and one \gls{stereocam} were estimated by comparing the fusion results for different values and choosing the ones providing the best performance.
We demonstrated the advantages of our approach compared to traditional combination rules such as the one's by Dempster and Yager based on real sensor data.
Future work will focus on combing the fused evidential grid maps from different time points in a recursive estimator to accumulated measurements over time and infer the dynamic state.

\printbibliography%

\end{document}